%% file: draft.tex
\documentclass[11pt]{article}

\usepackage{fullpage}
\usepackage{url}
\usepackage{smile}
\usepackage{pdfpages}
\usepackage{kpfonts}
\usepackage{tgpagella}
\usepackage{natbib}
\usepackage[]{hyperref}
\usepackage[left]{lineno}
\usepackage{blindtext}
\usepackage{wrapfig}
\usepackage[ruled,vlined,algo2e]{algorithm2e}
\usepackage{subfigure}
\newcommand\figcaption{\def\@captype{figure}\caption}

\usepackage{appendix}
\usepackage{microtype}
\usepackage{graphicx}
\usepackage{subfigure}
\usepackage{booktabs} 

\DeclareMathAlphabet{\mathsf}{OT1}{cmss}{m}{n}
\SetMathAlphabet{\mathsf}{bold}{OT1}{cmss}{bx}{n}

\makeatletter

\newcommand{\Rmnum}[1]{\expandafter\@slowromancap\romannumeral #1@}
\newcolumntype{P}[1]{>{\centering\arraybackslash}p{#1}}
\makeatother

\DeclareMathOperator*{\expect}{\EE}
\begin{document}


\title{\Huge Learning to Defend by Learning to Attack}

\date{\today}

\author{Haoming Jiang$^*$, Zhehui Chen$^*$, Yuyang Shi, Bo Dai, Tuo Zhao}

\maketitle

\begin{abstract}
Adversarial training provides a principled approach for training robust neural networks. From an optimization perspective, adversarial training is essentially solving a bilevel optimization problem. The leader problem is trying to learn a robust classifier, while the follower problem is trying to generate adversarial samples. Unfortunately, such a bilevel problem is difficult to solve due to its highly complicated structure. This work proposes a new adversarial training method based on a generic learning-to-learn (L2L) framework. Specifically, instead of applying existing hand-designed algorithms for the inner problem, we learn an optimizer, which is parametrized as a convolutional neural network. At the same time, a robust classifier is learned to defense the adversarial attack generated by the learned optimizer. Experiments over CIFAR-10 and CIFAR-100 datasets demonstrate that L2L outperforms existing adversarial training methods in both classification accuracy and computational efficiency. Moreover, our L2L framework can be extended to generative adversarial imitation learning and stabilize the training.
\end{abstract}

\input{intro}

\input{method}

\input{experiment}
\input{extension}
\input{discussion}

\bibliography{ref}
\bibliographystyle{ims}

\input{appendix_limitcycle}
\input{appendix}
\input{appendix_advcheck}

\end{document}

%% file: intro.tex

\section{Introduction}

This decade has witnessed great breakthroughs in deep learning in a variety of applications, such as computer vision~\citep{taigman2014deepface,girshick2014rich,he2016deep,liu2017deep}. Recent studies \citep{szegedy2013intriguing}, however, show that most of these deep learning models are very vulnerable to adversarial attacks. Specifically, by injecting a small perturbation to a normal sample, one can obtain an adversarial sample. Although the adversarial sample is semantically indistinguishable from the normal one, it can fool deep learning models and undermine the security of deep learning, causing reliability problems in autonomous driving, biometric authentication, etc. 

Researchers have devoted many efforts to study efficient adversarial attack and defense \citep{szegedy2013intriguing,goodfellow2014explaining,nguyen2015deep,zheng2016improving,madry2017towards,carlini2017towards}. There is a growing body of work on generating adversarial samples, such as, fast gradient sign method (FGSM, \cite{goodfellow2014explaining}), projected gradient method (PGM, \cite{kurakin2016adversarial}), Carlini-Wagner (CW,~\cite{paszke2017automatic}), etc. As for defense, existing methods can be unified as a bilevel optimization problem as follows: 
\begin{align}
\textrm{(Leader)}~  &\quad\quad\textstyle\min_{\boldsymbol{\theta}}~~ \expect_{P^*}\big[\ell(f_{\boldsymbol{\theta}}(\tilde{\bx}),\tilde{\by})\big],\label{eqn:upper}\\
\textrm{(Follower)}~  &\textrm{s.t.}~\textstyle P^* \in\argmax\limits_{\tilde{P}\in\cP} \expect_{\tilde{P}}\big[q_{f_{\boldsymbol{\theta}}}\big((\bx,\by),(\tilde{\bx},\tilde{\by})\big)\big],\notag
\end{align}
where $\ell$ denotes the loss function, $f_{\boldsymbol{\theta}}$ denotes the neural network classifier with parameter $\boldsymbol{\theta}$, $(\bx,\by)$ denotes the clean sample from distribution $D$,  $q_{f_{\boldsymbol{\theta}}}(\cdot,\cdot)$ denotes a measure depending on network $f_{\boldsymbol{\theta}}$, and $\cP$ denotes a set of joint distributions of perturbed sample $(\tilde{\bx},\tilde{\by})$ and clean sample $(\bx,\by)$. Here $\tilde{P}\in\cP$ satisfies that in each sample $(\tilde{\bx},\tilde{\by})$ is close to $(\bx,\by)$ and the marginal distribution of $\tilde{P}$ over $(\bx,\by)$ is $D$. By solving the follower problem in \eqref{eqn:upper}, $P^*$ essentially represents an effective adversarial distribution. Existing adversarial training methods use different approaches to find $P^*$ under different $q_{f_{\boldsymbol{\theta}}}$ and $\cP$. For example, \cite{goodfellow2014explaining} consider a special case of this problem, distributionally robust optimization (DRO, \cite{gao2016distributionally,rahimian2019distributionally}). In DRO, $q_{f_{\boldsymbol{\theta}}}$ in \eqref{eqn:upper} is the same as $\ell$ in \eqref{eqn:upper} and $\tilde{P}\in\cP$ satisfies that in each sample $\tilde{\by}=\by$, i.e., train the network $f_{\boldsymbol{\theta}}$ over adversarial samples and still require $f_{\boldsymbol{\theta}}$ to yield the correct labels. Another example is adversarial interpolation training (AIT, \cite{zhang2019adversarial}), where $q_{f_{\boldsymbol{\theta}}}$ is the cosine similarity between the features of adversarial sample and clean sample, and $\cP$ is a set of adversarial distribution yielded by mixup \cite{zhang2017mixup}. More details are in Section~\ref{sec:pre}. 

\eqref{eqn:upper} contains two optimization problems, referred to as leader and follower problems, respectively in the optimization literature. Such a  bilevel formulation naturally provides us a unified perspective on prior works of robustifying the neural network: The leader aims to find a robust network  so that the loss given by the training distribution from the follower problem is minimized; The follower targets on finding an optimal distribution that maximizes a certain measure, which yields a distribution of adversarial samples. 

Though the bilevel problem is straightforward and well formulated, it is hard to solve. Even the simplest version of bilevel problem, linear-linear bilevel optimization, is shown to be NP-hard~\citep{colson2007overview}. In our case, the problem becomes more challenging, since loss function $\ell$ in the leader is highly nonconvex in $\boldsymbol{\theta}$ and the follower targets on finding an optimal distribution under a nonconcave measure $q_{f_{\boldsymbol{\theta}}}$. Besides, the feasible domain of the follower problem is a space of continuous distributions; while, in practice, we have finite samples to approximate the original problem. Such a gap makes the problem more challenging.

There are several approaches to solve the original problem~\eqref{eqn:upper}.  Under the DRO setting, \cite{goodfellow2014explaining} propose to use FGSM to solve the DRO. However, \cite{kurakin2016adversarial} then find that FGSM with true label suffers from a ``label leaking'' issue, which ruins adversarial training. \cite{madry2017towards} further suggest to find adversarial samples by PGM and outperforms FGSM, since FGSM essentially is one iteration PGM; Alternatively, \cite{zhang2019adversarial} propose to combine FGSM and mixup to yield an adversarial samples for both feature and label. All these methods need to find an adversarial $(\tilde{\bx}_i,\tilde{\by}_i)$ for each clean sample $(\bx_i,\by_i)$, thus the dimension of the overall search space for all samples is substantial, which makes the computation expensive. More recently, \cite{li2019nattack} propose to use the natural evolution strategy to learn an adversarial distribution under the black-box setting, which is beyond the scope of this paper. 

To address the above challenges, we propose a new learning-to-learn (L2L) framework that provides a more principled and efficient way for solving adversarial training. Specifically, we parameterize the optimizer of the follower problem by a neural network denoted by $g_{\boldsymbol{\phi}}(\cA_{f_{\boldsymbol{\theta}}}(\bx,\by))$, where $\cA_{f_{\boldsymbol{\theta}}}(\bx,\by)$ denotes the input of the optimizer $g_{\boldsymbol{\phi}}$ with parameter $\boldsymbol{\phi}$. We also call the optimizer as the attacker. Since the neural network is very powerful in function approximation, our parameterization ensures that $g_{\boldsymbol{\phi}}$ is able to yield strong adversarial samples. Under our framework, instead of directly solving the follower problem in \eqref{eqn:upper}, we update the parameter $\boldsymbol{\phi}$ of the optimizer $g_{\boldsymbol{\phi}}$. Our training procedure becomes updating the parameters of two neural networks, which is quite similar to generative adversarial network~(GAN, \cite{goodfellow2014generative}). The proposed L2L is a generic framework and can be extended to other bilevel optimization problems, e.g., generative adversarial imitation learning, which is studied in Section~\ref{app:extention}.

Different from the hand-designed methods that compute adversarial perturbation $\boldsymbol{\delta}_i=\tilde{\bx_i}-\bx_i$ for each individual sample $(\bx_i,\by_i)$ using gradients from backpropagation, our methods generate perturbations for all samples through the shared optimizer $g_{\boldsymbol{\phi}}$. This enables optimizer $g_{\boldsymbol{\phi}}$ to learn potential common structures of the perturbations. Therefore, our method is capable of yielding strong perturbations and accelerating the training process. Furthermore, the L2L framework is very flexible: we can either choose different input $\cA_{f_{\boldsymbol{\theta}}}(\bx,\by)$, or use different architecture. For example, we can include gradient information in $\cA_{f_{\boldsymbol{\theta}}}(\bx,\by)$ and use a recurrent neural network (RNN) to mimic multi-step gradient-type methods. Instead of computing the high order information with finite difference approximation or multiple gradients, by parameterizing the algorithm as a neural network, our proposed method can capture this information in a much adaptive way~\citep{finn2017model}. Our experiments demonstrate that L2L not only outperforms existing adversarial training methods, e.g., PGM training, but also enjoys computational efficiency over CIFAR-10 and CIFAR-100 datasets~\citep{krizhevsky2009learning}.

The research on L2L has a long history~\citep{schmidhuber1987evolutionary,schmidhuber1992learning,schmidhuber1993neural,younger2001meta,hochreiter2001learning,andrychowicz2016learning}. The basic idea is that the updating formula of complicated optimization algorithms is first modeled in a parametric form, and then parameters are learned by some simple algorithms, e.g., stochastic gradient algorithm. Among existing works, \cite{hochreiter2001learning} propose a system allowing the output of backpropagation from one network to feed into an additional learning network, with both networks trained jointly; \cite{andrychowicz2016learning} then show that the design of an optimization algorithm can be cast as a learning problem. Specifically, they use long short-term memory RNNs to model the algorithm and allow the RNNs to exploit structure in the problems of interest in an adaptive way, which is one of the most popular methods for L2L. 

However, there are two major drawbacks of the existing L2L methods: {\bf (1)} It requires a large amount of datasets (or a large number of tasks in multi-task learning) to guarantee the learned optimizer to generalize, which limits their applicability (most of the related works only consider the image encoding as the motivating application); {\bf (2)} The number of layers/iterations in RNNs for modeling algorithms cannot be large to avoid computational burden.

Our contribution is that we fill the blank of L2L framework in solving bilevel optimization problems, and our proposed methods do not suffer from the aforementioned drawbacks: {\bf (1)} Different $f_{\boldsymbol{\theta}}$ and $(\bx,\by)$ yield different follower problems. Therefore, for adversarial training, we have sufficiently many tasks for L2L; {\bf (2)} The follower problem does not need a large scale RNN, and we use a convolutional neural network (CNN) or a length-two RNN (sequence of length equals 2) as our attacker network, which eases computation. Our code is available at \url{https://github.com/YuyangShi/Learning-to-Defend-by-Learning-to-Attack}.

\noindent \textbf{Notations}. Given a scalar $a\in \RR$, denote $(a)_+$ as $\max(a,0)$. Given two vectors $\bx,\by\in\RR^d$, denote $x_i$ as the $i$-th element of $\bx$, $\norm{\bx}_{\infty}=\max_i|x_i|$ as the $\ell_{\infty}$-norm of $\bx$, $\bx \circ \by=[x_1 y_1,\cdots,x_d y_d]^\top$ as element-wise product, and $\be_i$ is the vector with $i$-th element as $1$ and others as $0$. Denote the simplex in $\RR^d$ by $\Delta(d): =\{\bx: \norm{\bx}_1 = 1\}$, the $\ell_\infty$-ball centered at $\bx$ with radius $\epsilon$ by $\cB(\bx,\epsilon) = \{\by\in\RR^d: \norm{\by-\bx}_{\infty} \leq \epsilon\}$ and the projection to $\cB(\mathbf{0},\epsilon)$ as $\Pi_{\epsilon}(\boldsymbol{\delta})=\sign(\boldsymbol{\delta})\circ\max(|\boldsymbol{\delta}|,\epsilon)$, where $\sign$ and $\max$ are element-wise operators.

%% file: method.tex

\section{Preliminary}\label{sec:pre}

We focus on the defense against $\ell_\infty$-norm attack. In this section, we first introduce two popular cases of the original problem: distributionally robust optimization (DRO) and adversarial interpolation training (AIT). Then we discuss the fundamental hardness of solving these problems and the drawbacks of existing methods. 

\subsection{Adversarial Training}

Instead of using population loss in \eqref{eqn:upper}, we use empirical loss in the following context, since in practice we only have finite samples. Given $n$ samples $\{(\bx_i,\by_i)\}_{i=1}^n$, where $\bx_i$ is the $i$-th image and $\by_i$ is the corresponding label, DRO aims to solve:
\begin{align}
&\hspace{0.3in}\min_{\boldsymbol{\theta}}~~ \frac{1}{n}\sum_{i=1}^n \big[\ell(f_{\boldsymbol{\theta}}(\bx_i+{\boldsymbol{\delta}}_i),\by_i)\big],\label{eqn:dro_up}\\
& \textrm{s.t.}\quad {\boldsymbol{\delta}}_i\in \argmax_{\boldsymbol{\delta}\in\cB(\mathbf{0},\epsilon)}\ell(f_{\boldsymbol{\theta}}(\bx_i+\boldsymbol{\delta}),\by_i).\label{eqn:dro_low}
\end{align}
The standard pipeline of DRO version is shown in Algorithm~\ref{alg:advtrain}.  Since the step of generating adversarial perturbation $\boldsymbol{\delta}_i$ in Algorithm~\ref{alg:advtrain} is intractable, most adversarial training methods adopt hand-designed algorithms. For example, \cite{kurakin2016adversarial} propose to solve follower problem~\eqref{eqn:dro_low} approximately by first order methods like PGM. Specifically, PGM iteratively updates the adversarial perturbation by the projected sign gradient ascent method for each sample: Given sample $(\bx_i,\by_i)$, at the $t$-th iteration, PGM takes 
\begin{align}\label{eqn:pgm}
 \boldsymbol{\delta}_i^t \leftarrow \Pi_{\epsilon}\big(\boldsymbol{\delta}_i^{t-1}+\eta \cdot \sign\big(\nabla_{\bx} \ell(f_{\boldsymbol{\theta}}(\tilde{\bx}^{t}_i),\by_i)\big)\big),
\end{align}
where $\tilde{\bx}^{t}_i=\bx_i+\boldsymbol{\delta}_i^{t-1}$, $\eta$ is the perturbation step size, $T$ is a pre-defined total number of iterations, and $\boldsymbol{\delta}_i^0=\mathbf{0}$, $t=1,\cdots,T$. Finally PGM takes $\boldsymbol{\delta}_i = \boldsymbol{\delta}_i^T$. Note that FGSM essentially is one-iteration PGM. Besides, some works adopt other optimization methods, e.g., momentum gradient method \citep{dong2018boosting}, and L-BFGS~\citep{tabacof2016exploring}. 

\begin{algorithm}[H]
	\caption{\it Distributionally Robust Optimization.}\label{alg:advtrain}
		\textbf{Input: }$\{(\bx_i,\by_i)\}_{i=1}^n$: data, $\alpha$: learning rate, $N$: number of iterations, $\epsilon$: perturbation magnitude. \\
		\For{\textrm{ $t$  $\gets 1$ to $N$ } }{
		{ Sample a minibatch $\cM_t$\\}
		\For{\textrm{ $i$  in $\cM_t$ }}
		{
		$\boldsymbol{\delta}_i\leftarrow \argmax_{\boldsymbol{\delta}\in\cB(\mathbf{0},\epsilon)} \ell(f_{\boldsymbol{\theta}}(\bx_i+\boldsymbol{\delta}),\by_i)\quad$ \hfill $\slash\slash$ Generate adversarial data.}
		$\boldsymbol{\theta}\leftarrow \boldsymbol{\theta}- \alpha \frac{1}{|\cM_t|}\sum_{i\in\cM_t}\nabla_{\boldsymbol{\theta}} \ell (f_{\boldsymbol{\theta}}(\bx_i+\boldsymbol{\delta}_i),\tilde{\by}_i)$\hfill $\slash\slash$Update $\boldsymbol{\theta}$ over adversarial data.
		}
\setlength{\textfloatsep}{0pt}
\end{algorithm}
Alternatively, AIT adopts the mixup method to generate an adversarial distribution for a given sample $(\bx_i,\by_i)$ and then randomly select a sample $(\tilde{\bx}_i,\tilde{\by}_i)$ from this adversarial distribution. Specifically, AIT solves the following problem:
\begin{align}
\min_{\boldsymbol{\theta}}~~ \frac{1}{n}\sum_{i=1}^n \EE_{(\tilde{\bx}_i,\tilde{\by}_i)\sim D_i}\big[\ell(f_{\boldsymbol{\theta}}(\tilde{\bx}_i),\tilde{\by}_i)\big],
\end{align}
where $D_i = \{(\tilde{\bx}^j_i,\tilde{\by}^j_i)\}_{j=1}^n$ is generated as follows:
\begin{align}
&\tilde{\bx}^j_i=\argmin_{\tilde{\bx}\in\cB(\bx_i,\epsilon) }\frac{f^s_{\boldsymbol{\theta}}(\bx_j)\cdot f_{\boldsymbol{\theta}}^s(\tilde{\bx})}{\norm{f^s_{\boldsymbol{\theta}}(\bx_j)}_2\norm{f^s_{\boldsymbol{\theta}}(\tilde{\bx})}_2},\quad \textrm{and} \quad \tilde{\by}^j_i = \argmin_{\tilde{\by}\in\Delta(C)\cap \cB(\by_i,\epsilon_{\by})}  \norm{\tilde{\by}-\frac{\mathbf{1} - \by_j}{C-1}}_2^2,\label{eqn:mixup_low}
\end{align}
where  $f_{\boldsymbol{\theta}}^s(\cdot)$ denotes the output of the $s$-th layer of network $f_{\boldsymbol{\theta}}$, $C$ denotes the number of classes, and $\mathbf{1}$ denotes the vector with all elements $1$. The standard pipeline is shown in Algorithm~\ref{alg:ait}. To ease the computation, \cite{zhang2019adversarial} use one-step gradient update as the solution of \eqref{eqn:mixup_low}. 

\subsection{Hardness}

Now we present the hardness for solving these problems. Ideally, we want to obtain the optima for the follower problem, i.e., 
\begin{align*}
  P^* :=\argmax_{\tilde{P}\in\cP} \expect_{\tilde{P}}\big[q_{f_{\boldsymbol{\theta}}}\big((\bx,\by),(\tilde{\bx},\tilde{\by})\big)\big].
\end{align*}
However, the measure $q_{f_{\boldsymbol{\theta}}}$ depends on network $f_{\boldsymbol{\theta}}$, which makes solving $P^*$ intractable. Therefore, in reality the sample $(\tilde{\bx}_i,\tilde{\by}_i)$ from the  obtained solution $\tilde{P}$ is very unlikely to be the sample $(\bx^*_i,\by^*_i)$ from $P^*$. This then often leads to a highly unreliable or even completely wrong search direction, i.e.,
\begin{align*}
 \langle \nabla_{\boldsymbol{\theta}} \ell(f_{\boldsymbol{\theta}}(\tilde{\bx}_i),\tilde{\by}_i), \nabla_{\boldsymbol{\theta}}\ell(f_{\boldsymbol{\theta}}(\bx_i^*),\by_i^*)\rangle<0,
\end{align*}
which may further result in a limiting cycle (See Appendix~\ref{app:limitcyc}). This becomes even worse when sample noises exist. Moreover, among the methods mentioned earlier, except FGSM, all require numerous queries for gradients, which is computationally expensive. 

\begin{algorithm}[H]
	\caption{\it Adversarial Interpolation Training.}\label{alg:ait}
		\textbf{Input: }$\{(\bx_i,\by_i)\}_{i=1}^n$: data, $\alpha$: learning rate, $N$: number of iterations, $\epsilon,\epsilon_{\by}$: perturbation magnitudes, $s$: the output layer of network, $C$: number of classes. \\
		\For{\textrm{ $t$  $\gets 1$ to $N$ } }{
		{ Sample a minibatch $\cM_t$\\}
		\For{\textrm{ $i$  in $\cM_t$ }}
		{
		Sample another index $j$ \\
		$\tilde{\by}_i \leftarrow (1-\epsilon_y) \by_i + \epsilon_y (\mathbf{1} - \by_j)/(C-1)$\\
		$\displaystyle\tilde{\bx}_i\leftarrow \argmin_{\tilde{\bx}\in\cB(\bx_i,\epsilon)}  \frac{f^s_{\boldsymbol{\theta}}(\bx_j)\cdot f^s_{\boldsymbol{\theta}}(\tilde{\bx})}{\norm{f^s_{\boldsymbol{\theta}}(\bx_j)}_2\norm{f^s_{\boldsymbol{\theta}}(\tilde{\bx})}_2} \quad$ \hfill$\slash\slash$ Generate adversarial data.}
		$\boldsymbol{\theta}\hspace{-0.025in} \leftarrow\hspace{-0.025in} \boldsymbol{\theta}\hspace{-0.025in} -\hspace{-0.025in} \alpha \frac{1}{|\cM_t|}\sum_{i\in\cM_t}\nabla_{\boldsymbol{\theta}} \ell (f_{\boldsymbol{\theta}}(\tilde{\bx}_i),\tilde{\by}_i)$ \hfill $\quad\slash\slash$ Update $\boldsymbol{\theta}$ over adversarial data.
		}
\end{algorithm}
\setlength{\textfloatsep}{0pt}

\section{Learning-to-Learn (L2L) Framework}\label{sec:method}
\begin{figure}[!htb]
\centering
\includegraphics[width = .85\textwidth]{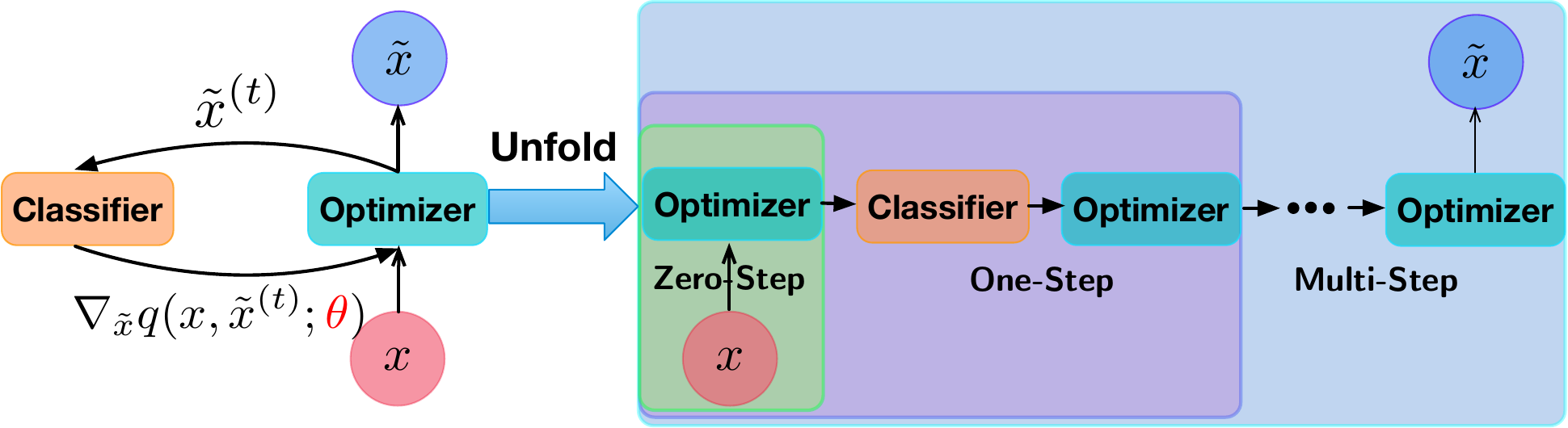}
\caption{\it An illustration of L2L: A neural network models optimizer for generating attack network. }\label{fig:idea}
\end{figure}

Since the hand-designed methods for bilevel problem~\eqref{eqn:upper} do not perform well, we propose to learn an optimizer for the follower problem. Specifically, we parameterize $\boldsymbol{\delta}=\tilde{\bx}-\bx$, the perturbation\footnote{This helps to handle the constraints $\boldsymbol{\delta}\in\cB(\mathbf{0},\epsilon)$.}, by a neural network $g_{\boldsymbol{\phi}}(\cA_{f_{\boldsymbol{\theta}}}(\bx,\by))$ with input $\cA_{f_{\boldsymbol{\theta}}}(\bx,\by)$ summarizing the information of data and classifier $f_{\boldsymbol{\theta}}(\cdot)$. We first show how our method works on the DRO: We convert DRO problem~\eqref{eqn:dro_up} and~\eqref{eqn:dro_low} to
\begin{align}\label{eqn:nn}
\min_{\boldsymbol{\theta}}\frac{1}{n}\sum_{i=1}^n\ell(f_{\boldsymbol{\theta}}(\bx_i+g_{\boldsymbol{\phi}}(\cA_{f_{\boldsymbol{\theta}}}(\bx_i,\by_i))),\by_i),
\end{align}
where $\boldsymbol{\phi}^*$ is defined as the solution to the problem:
\begin{align*}
 &\boldsymbol{\phi}^* \in \argmax_{\boldsymbol{\phi}} \frac{1}{n}\sum_{i=1}^n\ell(f_{\boldsymbol{\theta}}(\bx_i+g_{\boldsymbol{\phi}}(\cA_{f_{\boldsymbol{\theta}}}(\bx_i,\by_i))),\by_i),\\
 &~~~~\textrm{s.t. }\quad g_{\boldsymbol{\phi}}(\cA_{f_{\boldsymbol{\theta}}}(\bx_i,\by_i))\in \cB(\mathbf{0},\epsilon) , i \in [1,...,n].
\end{align*}
The optimizer $g_{\boldsymbol{\phi}}$ targets on generating optimal perturbations under constraints $g_{\boldsymbol{\phi}}(\cA_{f_{\boldsymbol{\theta}}}(\bx_i,\by_i))$ $\in\cB(\mathbf{0},\epsilon)$. These constraints can be handled by a $\mathrm{tanh}$ function and an $\epsilon$ scaler in the last layer of $g_{\boldsymbol{\phi}}$.
L2L framework is very flexible: We can choose different $\cA_{f_{\boldsymbol{\theta}}}(\bx,\by)$ as the input and mimic multi-step algorithms shown in Figure~\ref{fig:idea}. We provide three examples for DRO:

\noindent{\bf Naive Attacker.} This is the simplest example among our methods, taking original image $\bx_i$ as input, i.e., 
\begin{align*}
\cA_{f_{\boldsymbol{\theta}}}(\bx_i,\by_i)=\bx_i\quad \textrm{and} \quad \boldsymbol{\delta}_i=g_{\boldsymbol{\phi}}(\bx_i).
\end{align*} 
With this, L2L training is similar to GAN training. The major difference is that the generator in GAN yields synthetic data from random noises, while the naive attacker generates perturbations via samples.

\begin{figure}
	\centering
	\includegraphics[width=0.8\textwidth]{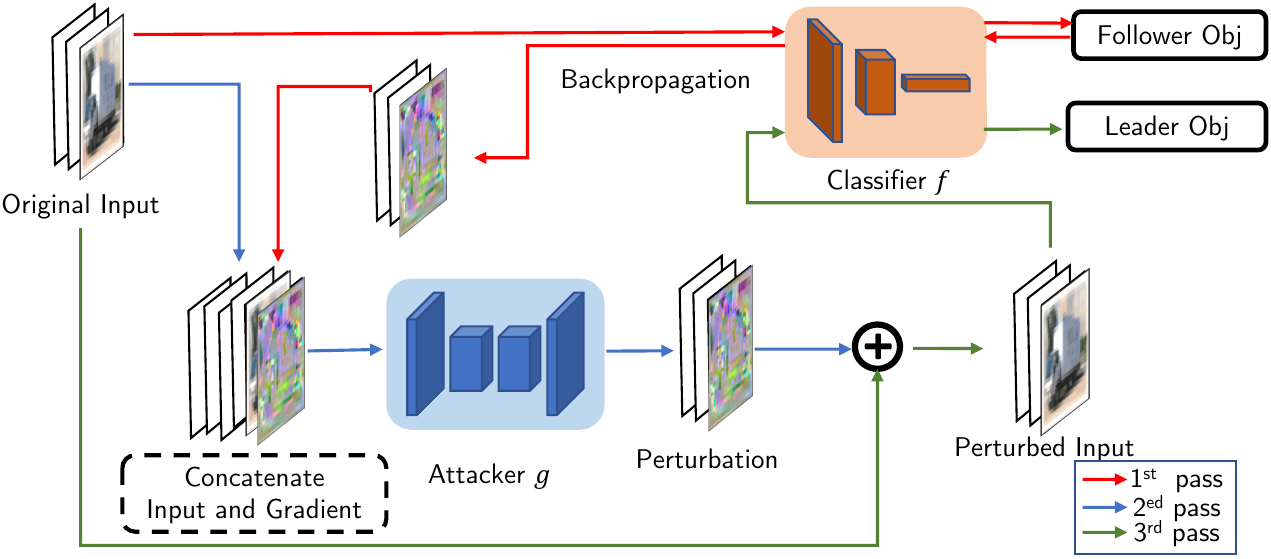}
	\caption{\it The architecture of adversarial training with gradient attacker model.}\label{fig:training}
\end{figure}

\noindent{\bf Gradient Attacker.} Motivated by FGSM, we design an attacker which takes the gradient information into consideration. Specifically, we concatenate image $\bx_i$ and gradient $\Delta_i = \nabla_{\bx} \ell(f_{\boldsymbol{\theta}}(\bx_i),\by_i)$ as the input of $g$:
\begin{align*}
\cA_{f_{\boldsymbol{\theta}}}(\bx_i,\by_i)=\big[\bx_i,\Delta_i \big]~~\textrm{and}~~ \boldsymbol{\delta}_i =g_{\boldsymbol{\phi}}\big([\bx_i,\Delta_i ]\big).
\end{align*}
With more information, the attacker is more effective to learn and yields more powerful perturbations.

\noindent{\bf Multi-Step Gradient Attacker.} Motivated by PGM, we adapt the RNN to mimic a multi-step gradient update. Specifically, we use the gradient optimizer network as the cell of RNN sharing the same parameter $\boldsymbol{\phi}$.  As we mentioned earlier, the number of layers/iterations in the RNN for modeling algorithms cannot be very large so as to avoid significant computational burden in backpropagation. In this paper, we focus on a length-two RNN to mimic a two-step gradient update. The corresponding perturbation becomes:
\begin{align*}
 \tilde{\bx}_i=\bx_i+ \Pi_{\epsilon}\big(\boldsymbol{\delta}_i^{(0)} + g_{\boldsymbol{\phi}} \big([\tilde{\bx}_i^{(0)},\nabla_{\bx} \ell(f_{\boldsymbol{\theta}}(\tilde{\bx}^{(0)}_i),\by_i) ]\big).
\end{align*}
Here $\boldsymbol{\delta}_i^{(0)} = g_{\boldsymbol{\phi}} \big([\bx_i,\nabla_{\bx} \ell(f_{\boldsymbol{\theta}}(\bx_i),\by_i)] \big)$, $\tilde{\bx}^{(0)}_i=\bx_i+\boldsymbol{\delta}_i^{(0)}$.
\begin{algorithm}[htb!]
	\caption{\it L2L-based DRO with gradient attacker.}\label{alg:l2l}
		\textbf{Input: }$\{(\bx_i,y_i)\}_{i=1}^n$: clean data, $\alpha_1,\alpha_2$: learning rates, $N$: number of epochs.\\
		\For{\textrm{$t$ $\gets$ $1$ to $N$}}{
	         Sample a minibatch $\cM_t$\\ 
                \For{\textrm{ $i$ in $\cM_t$ }}{
                             $\bu_i\leftarrow  \nabla_{\bx} \ell (f_{\boldsymbol{\theta}}(\bx_i),y_i),~~\boldsymbol{\delta}_i \leftarrow g_{\boldsymbol{\phi}}([\bx_i, \bu_i])$\hfill$\slash\slash$Generate perturbation by $g_{\boldsymbol{\phi}}$. 
		}
		$\boldsymbol{\theta} \leftarrow \boldsymbol{\theta} - \frac{ \alpha_1}{|\cM_t|}\sum\limits_{i\in\cM_t} \nabla_{\boldsymbol{\theta}} \ell (f_{\boldsymbol{\theta}}(\bx_i+\boldsymbol{\delta}_i),\by_i)$\hfill$\slash\slash$ Update $\boldsymbol{\theta}$ over adversarial data.\\
		 $\boldsymbol{\phi} \leftarrow \boldsymbol{\phi} + \frac{\alpha_2 }{|\cM_t|}\sum\limits_{i\in\cM_t} \nabla_{\boldsymbol{\phi}} \ell (f_{\boldsymbol{\theta}}(\bx_i+\boldsymbol{\delta}_i),\by_i)$\hfill$\slash\slash$ Update $\boldsymbol{\phi}$ over adversarial data.
		}
		\setlength{\textfloatsep}{0pt}
\end{algorithm}
Taking gradient attackers as an example, Figure~\ref{fig:training} illustrates how L2L works and jointly trains two networks: The first forward pass is used to obtain gradient of the classification loss over the clean data; The second forward pass is used to generate perturbation $\boldsymbol{\delta}_i$ by the attacker $g$; The third forward pass is used to calculate the adversarial loss $\ell$ in \eqref{eqn:nn}. Since our gradient attacker only needs one backpropagation, it amortizes the adversarial training cost, which leads to better computational efficiency. Moreover, L2L may adapt to the underlying optimization problem and yield better solution for the follower problem. The corresponding procedure of L2L is shown in Algorithm~\ref{alg:l2l}. 
	\begin{algorithm}[htb!]
		\caption{\it L2L-based AIT with gradient attacker.}\label{alg:l2l_ait}
		\textbf{Input: }$\{(\bx_i,\by_i)\}_{i=1}^n$: data, $\alpha_1$, $\alpha_2$: learning rates, $N$: number of iterations, $\epsilon_{\by}$: perturbation magnitudes. \\
		\For{\textrm{$t$ $\gets$ $1$ to $N$}}{
			Sample a minibatch $\cM_t$\\ 
			\For{\textrm{ $i$ in $\cM_t$ }}{
			Sample another index $j$\\
			$\tilde{\by}_i \leftarrow (1-\epsilon_{\by}) \by_i + \epsilon_{\by} (\mathbf{1} - \by_j)/(C-1)$, ~~ $\bu_i = \nabla_{\bx_i}q_{f_{\boldsymbol{\theta}}}(\bx_i,\bx_j)$, ~~$\boldsymbol{\delta}_i \leftarrow g_{\boldsymbol{\phi}}(\bx_i, \bu_i)$\\$\slash\slash$Generate perturbation by $g_{\boldsymbol{\phi}}$.
					}
			$\boldsymbol{\phi}\leftarrow \boldsymbol{\phi} -\frac{ \alpha_2}{|\cM_t|} \sum\limits_{i\in \cM_t}\nabla_{\boldsymbol{\phi}}q_{f_{\boldsymbol{\theta}}}(\bx_i+\boldsymbol{\delta}_i, \bx_j)$\hfill$\slash\slash$Update $\boldsymbol{\phi}$ over adversarial data.\\
			$\boldsymbol{\theta} \leftarrow \boldsymbol{\theta} - \frac{\alpha_1}{|\cM_t|}\sum\limits_{i\in\cM_t} \nabla_{\boldsymbol{\theta}} \ell (f_{\boldsymbol{\theta}}(\bx_i+\boldsymbol{\delta}_i),\tilde{\by}_i)$\hfill$\slash\slash$Update $\boldsymbol{\theta}$ over adversarial data.
		}
		\setlength{\textfloatsep}{0pt}
	\end{algorithm}

It is straightforward to extend L2L to AIT as shown in Algorithm~\ref{alg:l2l_ait}. We simply replace the gradient of $\ell$, $\nabla_{\bx} \ell(f_{\boldsymbol{\theta}}(\bx_i),\by_i)$, by the gradient of $q_{f_{\boldsymbol{\theta}}}(\bx_i,\bx_j)=\frac{f^s_{\boldsymbol{\theta}}(\bx_i)\cdot f_{\boldsymbol{\theta}}^s(\bx_j)}{\norm{f^s_{\boldsymbol{\theta}}(\bx_i)}_2\norm{f^s_{\boldsymbol{\theta}}(\bx_j)}_2}$, $\nabla_{\bx_i} q_{f_{\boldsymbol{\theta}}}(\bx_i,\bx_j)$ in the attacker input. Taking gradient network as an example, given a sample $(\bx_i,\by_i)$, we randomly select another sample $(\bx_j,\by_j)$, and yield the adversarial sample as follows:
\begin{align}
 \tilde{\bx}_i =\bx_i+ g_{\boldsymbol{\phi}}\Big([\bx_i,\nabla_{\bx_i} q_{f_{\boldsymbol{\theta}}}(\bx_i,\bx_j)]\Big),
\end{align}
and adopt the corresponding label vector $\tilde{\by}_i$ from \eqref{eqn:mixup_low}.

%% file: experiment.tex

\section{Experiments}\label{sec:exp}
	
To demonstrate the effectiveness and computational efficiency of L2L, we conduct experiments over both  CIFAR-10 and CIFAR-100 datasets. We compare our methods with original PGM training and adversarial interpolation training. All implementations are done in PyTorch with one single NVIDIA 2080 Ti GPU. Here we discuss the white-box setting, which is the most direct way to evaluate the robustness. 

\noindent{\bf Classifier Network.} All experiments adopt a 34-layer wide residual network (WRN-34-10, \cite{zagoruyko2016wide}) implemented by \cite{zhang2019theoretically} as the classifier network. For each method, we train the classifier network from scratch. 
\begin{table}[!htb]
\begin{center}
\caption{\it Attacker Architecture:  $k,c,s,p$ Denote the Kernel Size, Output Channels, Stride and Padding Parameters of Convolutional Layers, Respectively.}
\label{tab:optimizer}
{\small
\begin{tabular}{ l|l} 
				\hline
				\hline
				 Conv:& [$k = 3 \times 3, c=64, s = 1, p = 1$], BN+ReLU\\
				\hline
				  ResBlock:& [$k = 3 \times 3, c=128, s = 1, p = 1$] \\
				\hline
				ResBlock:& [$k = 3 \times 3, c=256, s = 1, p = 1$] \\
				\hline
				 ResBlock:& [$k = 3 \times 3, c=128, s = 1, p = 1$] \\
				\hline
				  Conv:& [$k = 3 \times 3, c=3, s = 1, p = 1$], $\mathrm{tanh}$\\
				\hline
				\hline
\end{tabular}}
\end{center}
\end{table}

\begin{figure}[!htb]
   \centering
\includegraphics[width = 0.7\textwidth]{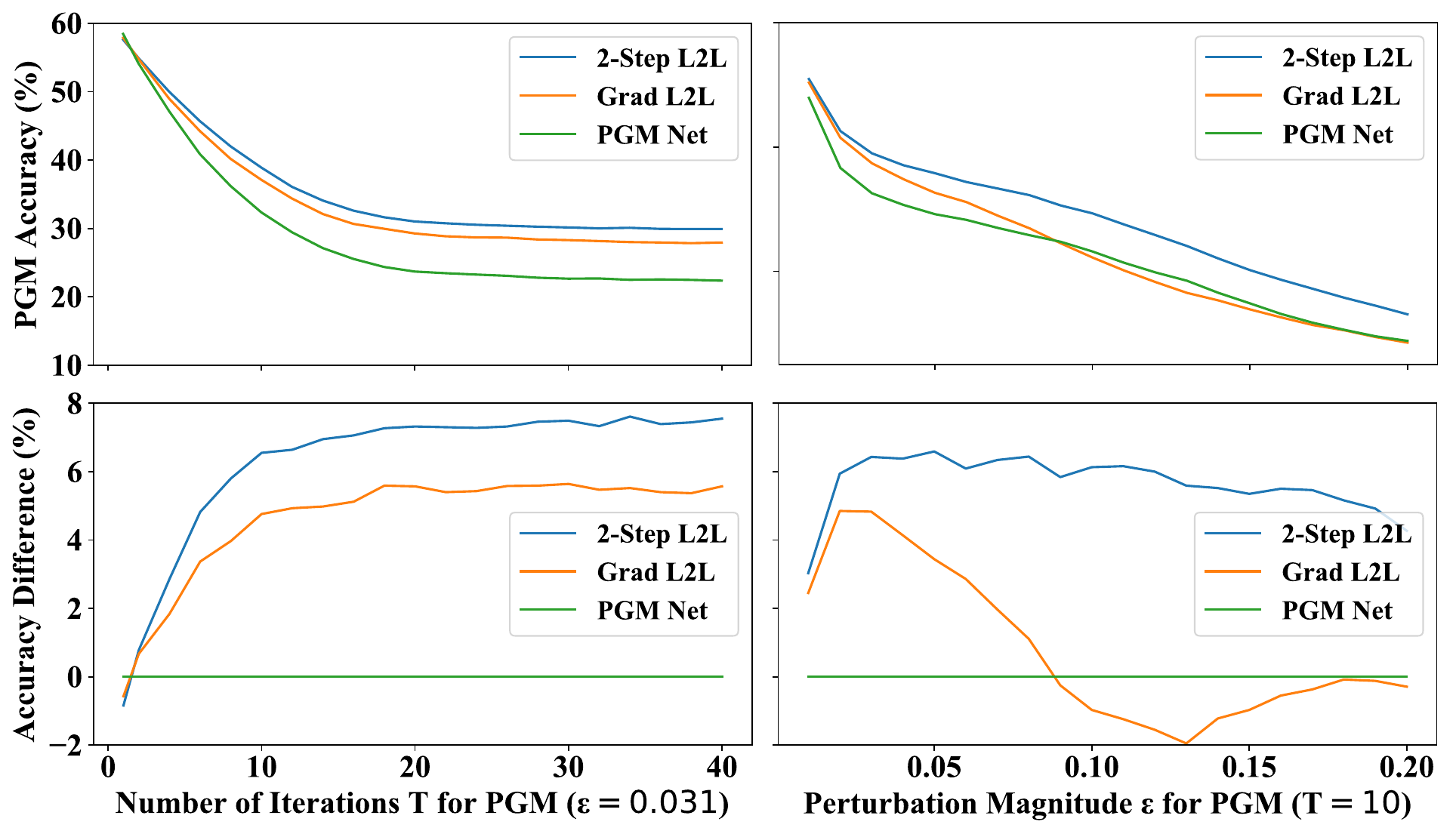}
   \caption{\it Robust accuracy against perturbation magnitude and number of iteration of PGM over CIFAR-100;. (Top) Accuracy; (Bottom) Performance gain over PGM Net. See more results in Appendix~\ref{app:advcheck}.}\label{fig:acc_vs_iter_eps}
\end{figure}

\noindent{\bf Attacker.}  Table~\ref{tab:optimizer} presents the architecture of our attacker network\footnote{We provide another attacker architecture  with down-sampling modules in the Section~\ref{sec:slim}. With such an attacker, L2L adversarial training is less stable, but faster. }. We adopt the ResBlock proposed in \cite{miyato2018spectral}. The detailed structure of ResBlock is provided in Appendix~\ref{sec:slim}. Batch normalization (BN) and activations, e.g., ReLU and $\mathrm{tanh}$, are applied when specified. The $\mathrm{tanh}$ function can easily make the output of attacker satisfy the constraints.

\begin{table*}[!htb]
	{\normalsize
		\begin{center}
			\caption{\it Results of Different Defense Methods under the White-box Setting.}
			\label{tab:cifar_white}
			\centering
			\begin{tabular}{c|c|c|c|c}
				\hline
				\hline 
				\multirow{2}{*}{Defense Method}  &  \multirow{2}{*}{Attack} & \multirow{2}{*}{Data Set}  & \multicolumn{2}{c}{Accuracy}\\
				\cline{4-5}
				& & &Clean & Robust
				\\
				\hline\hline \multicolumn{5}{l}{}  \\ [-1.8ex]
				\hline
				Stability Train \cite{zheng2016improving} &  PGM-20 & \multirow{2}{*}{CIFAR-10}      & 94.64\% & 0.15\% \\[0.005cm]
				PGM Net~\cite{madry2017towards} &  PGM-20 &      & 87.30\% & 47.04\% \\[0.005cm]
				\hline
				Naive L2L &    PGM-20     & \multirow{11}{*}{CIFAR-10}  & 94.53\% & 0.01\% \\[0.005cm]
				Grad-only L2L &    PGM-20    &   & 86.28\%&49.94\% \\[0.005cm]
				2-Step Grad-only L2L  &    PGM-20    &  & 85.8\%&53.85\%\\[0.005cm]
				Grad L2L &    PGM-20    &   & 85.84\%&51.17\% \\[0.005cm]
				2-Step L2L  &    PGM-20    &  &  85.35\%&54.32\%\\[0.005cm]
				Grad L2L	 &    PGM-100 &      &85.84\% &47.72\%\\[0.005cm]
				2-Step L2L&    PGM-100  &    &85.35\%&52.12\%\\[0.005cm]
				Grad L2L 	  &    CW &     &85.84\% &53.5\% \\[0.005cm]
				2-Step L2L&    CW   &   &85.35\% & 57.07\%\\[0.005cm]
				Grad L2L&    Random   &   &85.84\% & 82.67\%\\[0.005cm]
				2-Step L2L&    Random  &    &85.35\% & 83.10\%\\[0.005cm]
				Grad L2L&    Grad L2L   &   &85.84\% & 49.68\%\\[0.005cm]
				2-Step L2L&    2-Step L2L  &    &85.35\% & 52.71\%\\[0.005cm]
				\hline\hline \multicolumn{5}{l}{}  \\ [-1.8ex]
				\hline
				PGM Net &PGM-20 &  \multirow{9}{*}{CIFAR-100} &62.68\% & 23.75\% \\[0.005cm]
				Grad-only L2L &    PGM-20    &   & 62.4\%&27.64\%\\[0.005cm]
				2-Step Grad-only L2L  &    PGM-20    &  & 60.25\%&31.24\% \\[0.005cm]
				Grad L2L &    PGM-20   &    & 62.18\%&28.67\% \\[0.005cm]
				2-Step L2L  &    PGM-20 &     &  60.95\%&31.03\%\\[0.005cm]
				PGM Net&    PGM-100     &  & 62.68\%&22.06\%\\[0.005cm]
				Grad L2L &    PGM-100  &     &62.18\% &26.69\%\\[0.005cm]
				2-Step L2L&    PGM-100  &    &60.95\%&29.75\%\\[0.005cm]
				PGM Net  &    CW   &   & 62.68\%&25.95\% \\[0.005cm]
				Grad L2L  &    CW   &   & 62.18\%&29.65\% \\[0.005cm]
				2-Step L2L&    CW   &   & 60.95\%& 32.28\%\\[0.005cm]
				\hline
				\hline
			\end{tabular}
		\end{center}
	}
\end{table*}

\noindent{\bf White-box and Black-box.} We compare different methods under both white-box and black-box settings. Under the white-box setting, attackers can access all parameters of target models and generate adversarial examples based on the models; whereas under the black-box setting, we adopt the standard transfer attack method from~\cite{liu2016delving} as accessing parameters is prohibited. The results under the black-box setting are provided in Appendix~\ref{app:black}.

\noindent{\bf Robust Evaluation.} We evaluate the robustness of the networks by PGM and CW attacks with the maximum perturbation magnitude $\epsilon=0.031$ (after rescaling the pixels to $[0,1]$) over CIFAR 10 and 100. For PGM attack, we use 20 and 100-iteration PGM with a perturbation step size $\eta=0.003$, and for each sample we initialize the perturbation randomly in $\cB(0,10^{-4})$. For CW attack, we adopt the implementation in~\cite{paszke2017automatic}, and set the maximum number of iterations as $100$. For each method, we repeat $5$ runs with different random initial seed and report the {\it worst result}. %
For CIFAR-10, we also evaluate the robustness of Grad L2L and 2-Step L2L networks using random attacks, for which we uniformly sample $10^5$ perturbations in $\cB(0,0.031)$ adding to each test sample. We also evaluate the robustness of Grad L2L and 2-Step L2L networks under their own attackers. A full robustness checklist suggested by~\cite{carlini2019evaluating} is reported in Appendix~\ref{app:advcheck}.

\subsection{PGM Training}

For simplicity, we denote PGM Net as the classifier with PGM training, and Naive L2L, Grad L2L, and 2-Step L2L as the classifiers using L2L training with corresponding attackers. For reference, we also include some results of Grad-only L2L and 2-Step Grad-only L2L, whose attackers take the gradient information  only without the raw images.
	\begin{table*}[!htb]
	{\normalsize
		\begin{center}
			\centering
			\caption{\it One epoch running time. (Unit: s)}\label{tab:time}
			\centering
			\begin{tabular}{ P{.75in}P{.8in}P{.9in}P{.8in} P{.8in}P{.8in}@{}}
				\hline
				\hline
				Dataset&Plain Net &   PGM Net  & Naive L2L  & Grad L2L  & 2-Step L2L  \\
				\hline
				CIFAR-10 &$106.5 \pm 1.5$      & $1310.8 \pm 14.2$ & $293.7\pm 3.1$&$617.5\pm 6.1$& $805.1\pm 8.1$\\
				CIFAR-100 & $106.9 \pm 1.4$  & $1354.8 \pm 14.1$ & $310.0\pm 2.9$&  $623.1\pm 6.3$ & $824.7\pm 8.4$  \\
				\hline		
				\hline
			\end{tabular}	
	\end{center}}
\end{table*}

\noindent{\bf Original PGM.}  For CIFAR-10, we directly report the result from \cite{madry2017towards} as the baseline; For CIFAR-100, we train a PGM Net as the baseline: For optimizer, we use stochastic gradient descent (SGD) algorithm with Polyak's momentum (parameter $0.9$, \cite{liu2018toward}) and weight decay (parameter $2\times 10^{-4}$, \cite{krogh1992simple}). In addition, we adapt the setting from \cite{madry2017towards} but train the network for $100$ epochs with initial learning rate $0.1$, decay schedule [30,60,90], and decay rate $0.1$. For adversarial samples, we use a 10-iteration PGM with the perturbation step size $0.007$ in \eqref{eqn:pgm}.

\noindent{\bf PGM+L2L.} We train two networks for 100 epochs. For classifier's optimizer, we use the same configuration as original PGM training; For attacker's optimizer, we use Adam optimizer (parameter $[0.9,0.999]$, \cite{kingma2014adam}) with initial learning rate $10^{-3}$ (no learning rate decay) and weight decay (parameter $2\times 10^{-4}$) so that it adaptively balances the updates in both leader and follower optimization problems. 
\noindent{\bf Experiment Results.} Table~\ref{tab:cifar_white} shows the results of all PGM training methods over CIFAR-10 and 100 under the white-box setting. As can be seen, without gradient information, Naive L2L is vulnerable to the PGM attack. However, when the attacker utilizes the gradient information, Grad L2L and 2-Step L2L {\it significantly outperform} the PGM Net over CIFAR-10 and 100, with a slight loss for the clean accuracy. From the experiments on CIFAR-10, our Grad L2L and 2-Step L2L are robust to random attacks, where the accuracy is only slightly lower than the clean accuracy. Furthermore, the accuracy of our Grad/2-Step L2L model under the Grad/2-Step L2L attacker is comparable to the accuracy under PGM attacks, which shows that L2L attackers are able to generate strong attacks. As can be seen, PGM-100 is stronger than Grad L2L attacker ($47.72\%$ vs. $49.68\%$), but similar to the 2-Step L2L attacker ($52.07\%$ vs. $52.71\%$), which means 2-Step L2L attacker is much stronger than Grad L2L attacker and explains why  2-Step L2L is stronger than Grad L2L and PGM net. In addition, comparing Grad-only L2L with Grad L2L, we see that without the raw images fed into the attackers, Grad-only L2L is less robust to the PGM attack, though  2-Step Grad-only L2L and 2-Step L2L achieves comparable performance.

In addition, Table~\ref{tab:time} shows one epoch running time of all methods over CIFAR-10 and 100. As can be seen, Grad L2L and 2-Step L2L is much faster than PGM Net. By further comparing the accuracy of Grad/2-Step L2L and PGM Net in Table 2, we find that L2L methods enjoy computational efficiency. In addition, Figure~\ref{fig:acc_vs_iter_eps} presents the robust accuracy against number of iterations with $\epsilon = 0.031$ and perturbation magnitude (number of iterations $T=10$). As can be seen, 2-Step L2L is much more robust than PGM Net.

\subsection{Adversarial Interpolation Training}

We conduct the experiments of AIT over CIFAR-10 using the code from \cite{zhang2019adversarial}. \footnote{\url{https://github.com/Adv-Interp/adv_interp}}

\noindent{\bf Original AIT.} We follow the experimental setting in \cite{zhang2019adversarial}, but use a WRN-34-10. For classifier's optimizer, we use the same configuration in original PGM training. We choose the perturbation magnitude over label $\epsilon_{\by}$ as $0.5$. In addition, we train the whole network for 200 epochs with initial learning rate $0.1$, decay schedule [60,90], and decay rate $0.1$. Moreover, in each epoch, we first use FGSM to yield training samples via \eqref{eqn:mixup_low}, and then train the AIT Net over these adversarial samples. 

\noindent{\bf  AIT+L2L.} We train for 200 epochs. For classifier's optimizer, we adopt the configuration of SGD from the original AIT; For attacker's optimizer, we use Adam (parameter $[0.9,0.999]$) with initial step size as $10^{-3}$ (no decay) and weight decay (parameter $2\times 10^{-4}$).

\begin{table}[!htb]
	{\normalsize
		\begin{center}
			\caption{\it Results of AIT based defense methods under the white-box setting (CIFAR-10).}
			\label{tab:cifar_white_AIT}
			\centering
			\begin{tabular}{c|c|c|c}
				\hline
				\hline
				\multirow{2}{*}{Defense Method} &  \multirow{2}{*}{Attack}   & \multicolumn{2}{c}{Accuracy}\\
				\cline{3-4}
				& &Clean & Robust\\
				\hline
				AIT &    PGM-20       & 90.43\% & 75.33\% \\[0.005cm]
				Grad L2L &    PGM-20       & 91.65\%&80.87\% \\[0.005cm]
				AIT &    PGM-100       & 90.43\% & 67.84\% \\[0.005cm]
				 Grad L2L &    PGM-100       & 91.65\%&79.20\% \\[0.005cm]
				AIT &    CW-20       & 90.43\%& 64.79\%  \\[0.005cm] 
			        Grad L2L &    CW-20     & 91.65\% & 74.88\% \\[0.005cm]
				AIT &    CW-100       & 90.43\%&  61.69\% \\[0.005cm] 
				 Grad L2L &    CW-100       & 91.65\%&73.46\% \\[0.005cm]
				 
				\hline
				\hline
			\end{tabular}
		\end{center}
	}
\end{table}

\noindent{\bf Experiment Results.} Table~\ref{tab:cifar_white_AIT} shows the results of AIT methods over CIFAR-10 under the white-box setting. As can be seen, Grad L2L  {\it significantly improves upon} the AIT Net over CIFAR-10 on both clean accuracy and robust accuracy. 

\subsection{Visualization of Adversarial Examples} 

\begin{figure}[!htb]
	\centering
	\subfigure[PGM Net adv. samples]{
		\includegraphics[width=0.43\textwidth]{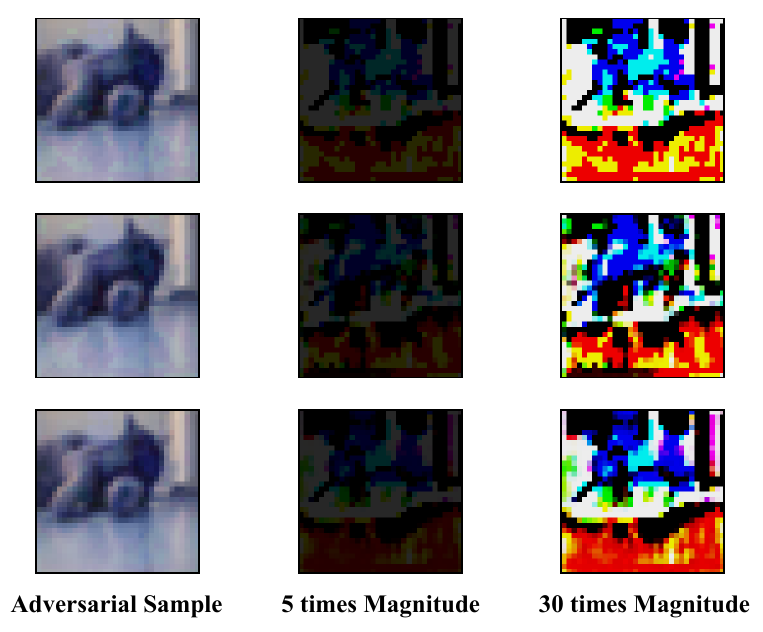}} 
	\subfigure[2-Step L2L adv samples]{
		\includegraphics[width=0.43\textwidth]{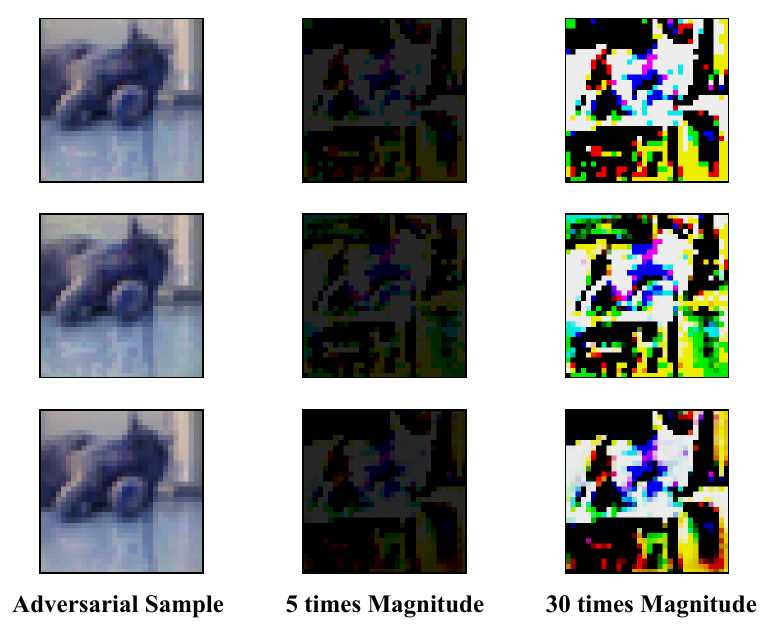}}
	\caption{\it Adv. examples of FGSM (Top), PGM-20 (Mid), 2-Step L2L (Bottom) perturbations for a cat under PGM Net and 2-Step L2L with $\epsilon=0.031$.}\label{perturbation}
\end{figure}

Figure~\ref{perturbation} provides an illustrative example of adversarial perturbations generated by FGSM, PGM-20 and 2-Step L2L attacker for a {\it cat} in CIFAR-10. As can be seen, attacks for these two networks are different. Moreover, the perturbation generated by the 2-Step L2L attacker is much smoother than FGSM and PGM. In this example, 2-Step L2L labels all adversarial samples correctly; whereas the PGM Net is fooled by PGM-20 attack and misclassifies it as a {\it dog}. 

\begin{figure}[!htb]
\centering
\subfigure[AIT adv. samples]{
\includegraphics[width=0.43\textwidth]{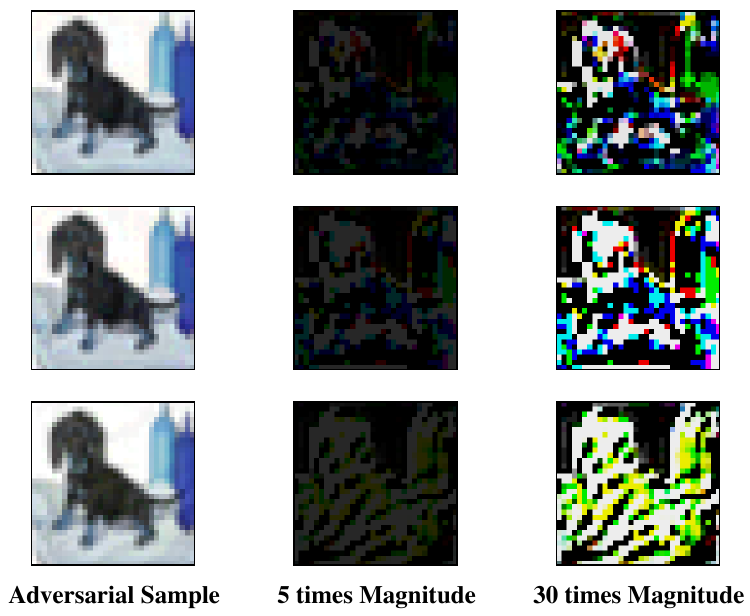}} 
\subfigure[Grad L2L adv. samples]{
\includegraphics[width=0.43\textwidth]{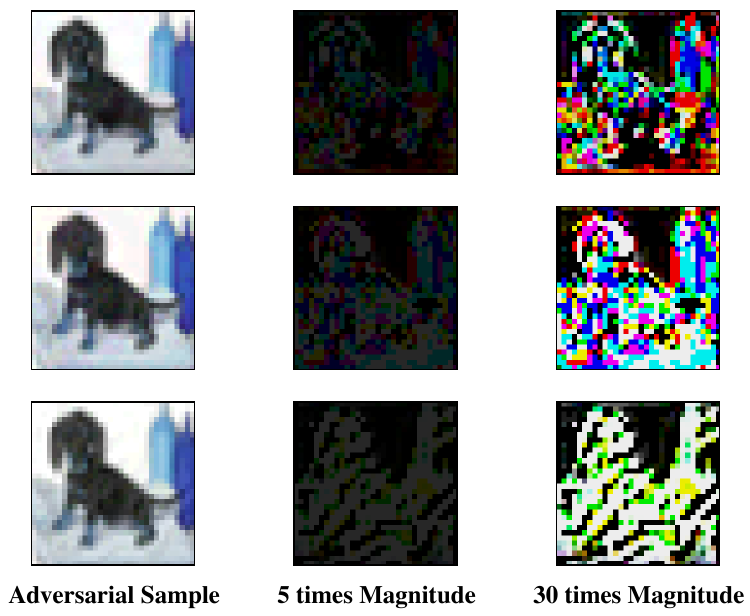}}
\caption{\it Illustrative adv. examples of PGM-20 (Top), AIT (Mid), and Grad L2L (Bottom) perturbations for a dog under AIT Net and Grad L2L with $\epsilon=0.031$.}
\label{perturbation_ait}
\end{figure}

Figure~\ref{perturbation_ait} provides an illustrative example of adversarial perturbations generated by PGM, AIT and Grad L2L for a {\it dog} in CIFAR-10. As can be seen, attacks for these two networks are very different: the attacks for the Grad L2L is more abundant in three channels. In this example, Grad L2L labels all adversarial samples correctly; whereas the AIT is fooled by all attacks and misclassifies it as a {\it horse}.

%% file: extension.tex
\section{Extension}\label{app:extention}

As we mentioned earlier that our proposed L2L framework is quite general, and applicable to a {\it broad} class of minimax optimization problems, here we present an extension of our proposed L2L framework to generative adversarial imitation learning (GAIL, \cite{DBLP:journals/corr/HoE16}) and conduct some numerical experiments for comparing the original GAIL and GAIL with L2L on two environments: CartPole and Mountain Car~\cite{brockman2016openai}.

\subsection{L2L for Generative Adversarial Imitation Learning}

Imitation learning aims to learn to perform a task from expert demonstrations, in which the learner is given only samples of trajectories from the expert. To solve this problem, GAIL tries to recover the expert's cost function and extract such a a policy from the recovered cost function, which can be formulated as the following bilevel optimization problem: 
\begin{align}
&\min_{\theta_\pi} L(\theta_\pi, \theta_{\rm D}^*) -\lambda H(\pi),\notag\\
\textrm{s. t. }~~&\theta_{\mathrm{D}}^*\in  \argmax_{{\theta_{\mathrm{D}}}} L(\theta_\pi, \theta_{\rm D}) + L_{\rm E}(\theta_{\rm D})-\lambda H(\pi),\label{eqn:gail} 
\end{align}
where $L(\theta_\pi, \theta_D) = \expect_{s,a \sim \pi(s;\theta_\pi)} [\log{(D(s,a;{\theta_{\mathrm{D}}}))}]$, $L_{\rm E}(\theta_{\rm D}) $$= \expect_{\tilde{s},\tilde{a} \sim \pi_{\mathrm{E}}}[\log (1-D(\tilde{s},\tilde{a};{\theta_{\mathrm{D}}}))]$ , $\pi(\cdot;\theta_\pi)$ is the trained policy parameterized by $\theta_\pi$, $\pi_{\mathrm{E}}$ denotes the expert policy, $D(\cdot,\cdot;{\theta_{\mathrm{D}}})$ is the discriminator parameterized by ${\theta_{\mathrm{D}}}$, $\lambda H(\pi)$ denotes a entropy regularizer with tuning parameter $\lambda$, $(s,a)$ and $(\tilde{s},\tilde{a})$ denote the state-action for the trained policy and expert policy, respectively. By optimizing \ref{eqn:gail}, the discriminator $D$ distinguishes the state-action $(s,a)$ generated from the learned policy $\pi$ with the sampled trajectories $(\tilde{s}, \tilde{a})$ generated from some expert policy $\pi_{\mathrm{E}}$. 
In the original GAIL training, for each iteration, we update the parameter of $D$, ${\theta_{\mathrm{D}}}$, by stochastic gradient ascend and then update $\theta_\pi$ by the trust region policy optimization (TRPO,~\cite{schulman2015trust}). 

Similar to the adversarial training with L2L, we apply our L2L framework to GAIL by parameterizing the inner optimizer as a neural network $U(; \theta_{\mathrm{U}})$ with parameter $\theta_{\mathrm{U}}$. Its input contains two parts: parameter $\theta_{\mathrm{D}}$ and the gradient of loss function with respect to $\theta_{\mathrm{D}}$: 
\begin{align*}
 g_{\mathrm{D}}(\theta_\mathrm{D},\theta_\pi) &=\expect\limits_{s,a \sim \pi(s;\theta_\pi)}[\nabla_{\theta_{\mathrm{D}}}\log{(D(s,a;\theta_{\mathrm{D}}))}] +\expect\limits_{\tilde{s},\tilde{a} \sim \pi_E} [\nabla_{\theta_{\mathrm{D}}}\log (1-D(\tilde{s},\tilde{a};\theta_{\mathrm{D}}))]. 
\end{align*}
In practice, we use a minibatch (several sample trajectories) to estimate $g_{\mathrm{D}}(\theta_{\mathrm{D}},\theta_\pi)$, denoted as $\hat{g}_{\mathrm{D}}(\theta_{\mathrm{D}},\theta_\pi).$ Specifically, at the $t$-th iteration, we first calculate $\hat{g}_{\mathrm{D}}^t = \hat{g}_{\mathrm{D}}(\theta_{\mathrm{D}}^t,\theta_\pi^t)$ and then update $\theta_{\mathrm{D}}^{t+1} = U( \theta_{\mathrm{D}}^t, \hat{g}_{\mathrm{D}}^t ; \theta_{\mathrm{U}}^t)$. Next, we update $\theta_{\mathrm{U}}$ by gradient ascend based on the sample estimate of 
\begin{align*}
&\expect\limits_{s,a \sim \pi(s;\theta^t_\pi)} [\nabla_{\theta_{\mathrm{U}}}\log{(D(s,a;\theta_{\mathrm{D}}^{t+1}))}] +\expect\limits_{\tilde{s},\tilde{a} \sim \pi_{\mathrm{E}}} [\nabla_{\theta_{\mathrm{U}}}\log (1-D(\tilde{s},\tilde{a};\theta_{\mathrm{D}}^{t+1}))].
\end{align*}
The detailed algorithm is presented in Algorithm~\ref{alg:gail}.

\begin{algorithm}[H]
	\caption{\it L2L-based GAIL.}\label{alg:gail}
	\textbf{Input: } $ \pi_E(\tilde{s})$: Expert;  $\theta_\pi$: Policy parameter;  $\theta_{\mathrm{D}}$: Discriminator parameter; $\theta_{\mathrm{U}}$: Updater parameter.\\
	\For{\textrm{$t$ $\gets$ $1$ to $N$}}{
		 $(s,a \sim \pi(a;\theta_\pi))$ $(\tilde{s},\tilde{a} \sim \pi_E(\tilde{s}))$\\
		$\slash \slash$ Sample trajectories and expert trajectories.
		\\
		$g^t_{\mathrm{D}} \leftarrow \frac{1}{|(s,a)|}\sum\limits_{(s,a)} [\nabla_{\theta_{\mathrm{D}}}\log{(D(s,a;\theta^t_{\mathrm{D}}))}] + \frac{1}{|(\tilde s,\tilde a)|}\sum\limits_{(\tilde s,\tilde a)} [\nabla_{\theta_{\mathrm{D}}}\log (1-D(\tilde{s},\tilde{a};\theta^t_{\mathrm{D}}))]$\\$\slash\slash$Compute gradient.\\
		$\theta_{\mathrm{D}}^{t+1} = U( \theta_{\mathrm{D}}^t, g_{\mathrm{D}}^t ; \theta_{\mathrm{U}}^t)$\\
		$\slash\slash$Update the discriminator parameters.  \\ 
		$\theta_{\mathrm{U}}^{t+1} \leftarrow \argmin\limits_{\theta_{\mathrm{U}}} \frac{1}{|(s,a)|}\sum\limits_{(s,a)} [\log{(D(s,a;\theta_{\mathrm{D}}^{t+1}))}]+\frac{1}{|(\tilde s,\tilde a)|}\sum\limits_{(\tilde s,\tilde a)} [\log (1-D(\tilde{s},\tilde{a};\theta_{\mathrm{D}}^{t+1}))]$ \\$\slash\slash$Update $\theta_{\mathrm{U}}$ of updater. \\
		Update $\theta_\pi$ by a policy step using the TRPO rule \cite{DBLP:journals/corr/HoE16}\\
		$\slash\slash$Update policy parameter $\theta_\pi$.
	}
	\setlength{\textfloatsep}{0pt}
\end{algorithm}

\subsection{Numerical Experiments}

\noindent \textbf{Updater Architecture.} We use a simple 3-layer perceptron with a skip layer as our updater. The number hidden units are ($2m \rightarrow 8m \rightarrow 4m \rightarrow m$), where $m$ is the dimension of $\theta_{\mathrm{D}}$ that depends on the original task. For the first and second layers, we use Parametric ReLU (PReLU, \cite{he2015delving}) as the activation function, while the last layer has no activation function. Finally we add the output to $\theta_{\mathrm{D}}$ in the original input as the updated parameter for the discriminator network.

\noindent{\textbf{Hyperparameter Settings.}} For all baselines we exactly follows the setting in \cite{DBLP:journals/corr/HoE16}, except that we use a 2-layer discriminator with number of hidden units ($(s,a) \rightarrow 64 \rightarrow 32 \rightarrow 1$) using $\tanh$ as the activation function. We use the same neural network architecture for $\pi$ and the same optimizer configuration. The expert trajectories are obtained by an expert trained using TRPO. For L2L based GAIL, we also use Adam optimizer to update the $\theta_{\mathrm{U}}$ with the same configuration as updating $\theta_{\mathrm{D}}$ in the original GAIL. 

\begin{figure*}[htb!]
	\centering
	\includegraphics[width=0.95\textwidth]{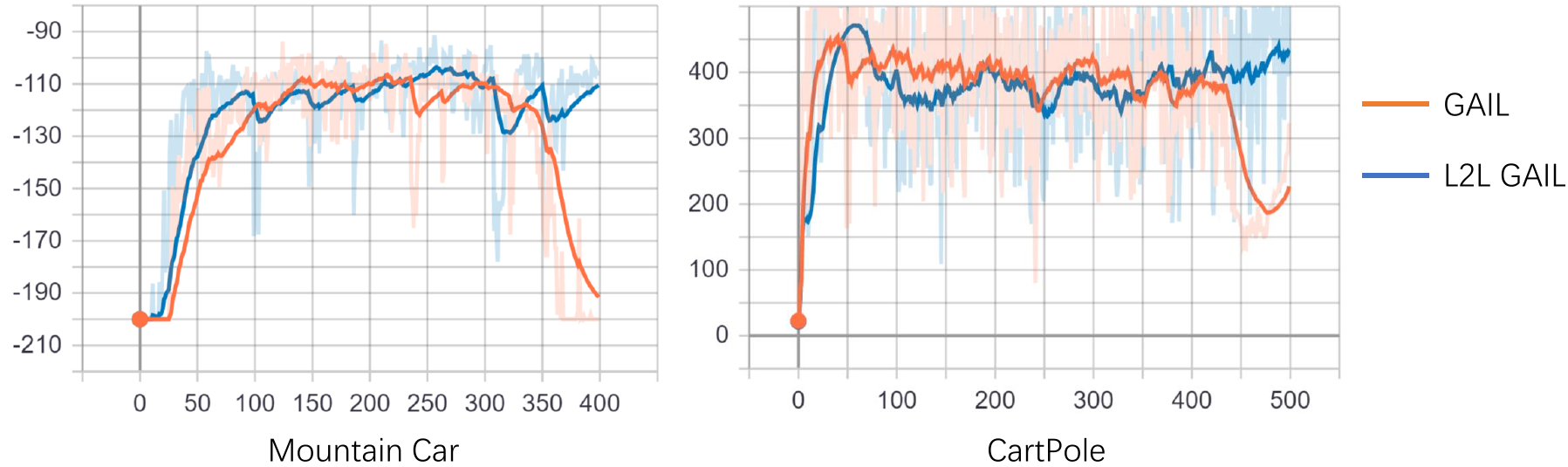}
	\caption{\it Reward vs. iteration of the trained policy using original GAIL and L2L GAIL under two environments: Mountain Car and CartPole.}
	\label{fig:gail}
\end{figure*}

\noindent{\textbf{Numerical Results.}} As can be seen in Figure~\ref{fig:gail}, GAIL has a sudden performance drop after training for a long time. We conjecture that this is because the discriminator overfits the expert trajectories and converges to a bad optimum, which is not generalizable. On the other hand, GAIL with L2L is much more stable. It is very important to real applications of GAIL: since the reward in real-world environment is usually unaccessible, we cannot know whether there is a sudden performance drop or not. With L2L, we can stabilize the training and obtain a much more reliable algorithm for real-world applications.

%% file: discussion.tex

	\vspace{-0.15in}
\section{Discussions}\label{sec:dis}

We discuss several closely related works: 

\noindent $\bullet$ By leveraging the Fenchel duality and feature embedding technique, \cite{dai2016learning} convert a learning conditional distribution problem to a minimax problem, which is similar to our naive attacker. Both approaches, however, lack the primal information. In contrast, gradient attacker network considers the gradient information of primal variables, and achieves good results with this key information. 

\noindent $\bullet$  \cite{goodfellow2014generative} propose the GAN, which is very similar to our L2L framework. Both GAN and L2L contain one generator network and one classifier network, and jointly train these two networks. There are two major difference between GAN and our framework: (1) GAN aims to transform the random noises to the synthetic data which is similar to the training examples, while ours targets on transforming the training examples to the adversarial examples for robustifying the classifier; (2) Our attacker does not only take the training examples (analogous to the random noise in GAN) as the input, but also exploits the gradient information of the objective function, since it essentially represents an optimization algorithm. The training procedure of these two, however, are quite similar. We adopt some tricks from GAN training to our framework to stabilize training process, e.g., in Grad L2L, we use the two-time scale trick~\citep{heusel2017gans}.

\noindent $\bullet$ There are some other works simply combining the GAN framework and adversarial training together. For example, \cite{baluja2017adversarial} and \cite{xiao2018generating} propose some ad hoc GAN-based methods to robustify neural networks. Specifically, for generating adversarial examples, they only take training examples as the input of the generator, which lacks the information of the outer mimnimization problem. Instead, our proposed L2L methods (e.g., Grad L2L, 2-step L2L) connect outer and inner problems by delivering the gradient information of the objective function to the generator. This is a very important reason for our performance gain on the benchmark datasets. As a result, the aforementioned GAN-based methods are only robust to simple attacks, e.g., FGSM, on simple data sets, e.g., MNIST, but fail for strong attacks, e.g., PGM and CW, on complicated data sets, e.g. CIFAR, where our L2L methods achieve significantly better performance. 

\noindent{\bf Training Stability}: For improving the training stability, we use both clean image and the corresponding gradient as the input of the attacker. Without such gradient information, the attacker severely suffers from training instability, e.g., the Naive Attacker Network. Furthermore, we try another architecture with the widely used downsampling modules, called ``slim attacker'' in Section~\ref{sec:slim}. We observed that the slim attacker also suffers from training instability. We suspect that the downsampling causes the loss of information. Thus, we tried to enhance the slim attacker by skip layer connections. In this way, the training is stabilized. However, the robust performance is still worse than the proposed architecture. 
 
\noindent{\bf Benefits of our L2L in adversarial training}:\\
{\bf (1)} Since neural networks have been known to be powerful in function approximation, our attacker $g$ can yield strong adversarial perturbations. Since they are generated by the same attacker, attacker $g$ learns some common structures across all samples; \\
{\bf (2)} {\it Overparametrization} is conjectured to ease the training of deep neural networks. We believe that similar phenomena happen to our attacker network, and ease the adversarial training.\\
%
%
%
%

%% file: appendix_limitcycle.tex
\newpage
\onecolumn
\appendices
\noindent\rule[0.5ex]{\linewidth}{4pt}
\begin{center}
\textbf{\LARGE Supplementary Materials}
\end{center}
\noindent\rule[0.5ex]{\linewidth}{1pt}

\section{Limiting Cycle}\label{app:limitcyc}

\begin{figure}[!htb]
\begin{center}
\includegraphics[width=.45\textwidth]{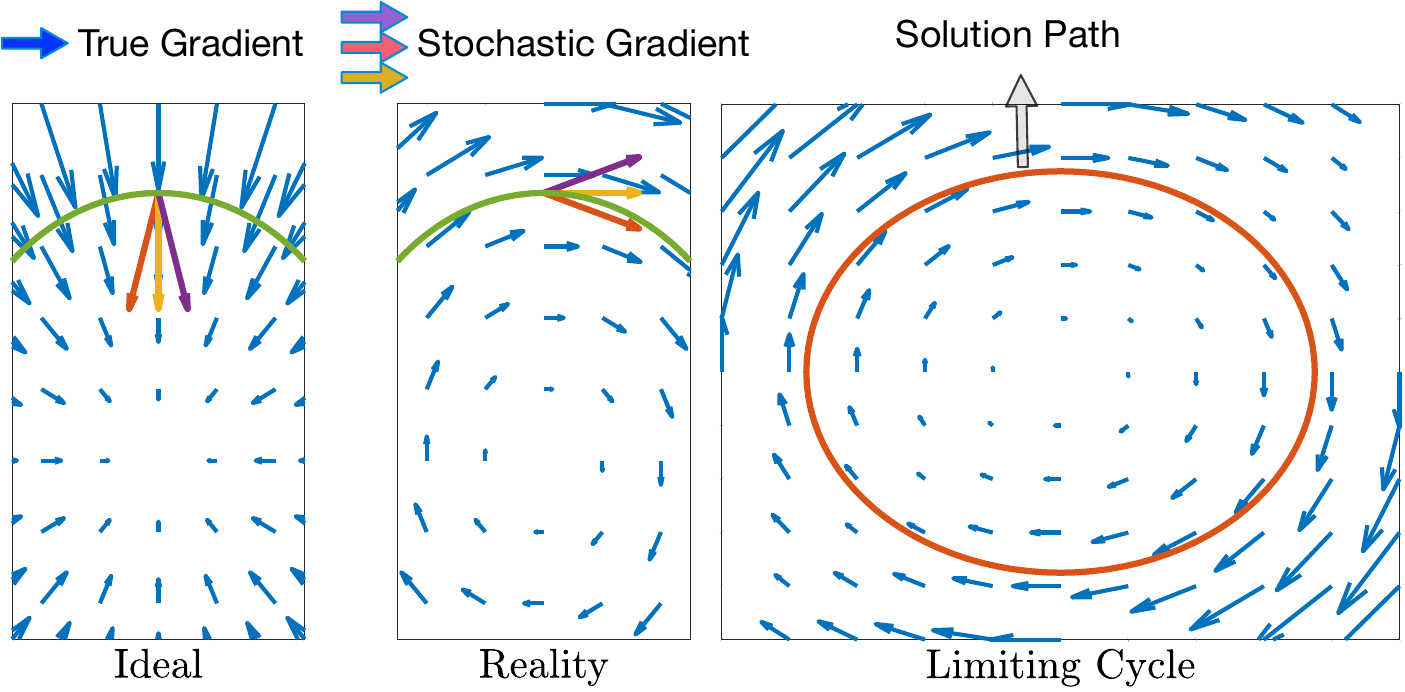}

\caption{\it Illustrative hardness for solving problem~\eqref{eqn:upper}. Wrong directions can lead to a limiting cycle. Then algorithms fail to converge. Details in Appendix~\ref{app:limitcyc}.}\label{Oscillation_Vis3}
\end{center}
\end{figure}

\begin{figure}[htb!]
\centering
\includegraphics[width=0.5\textwidth]{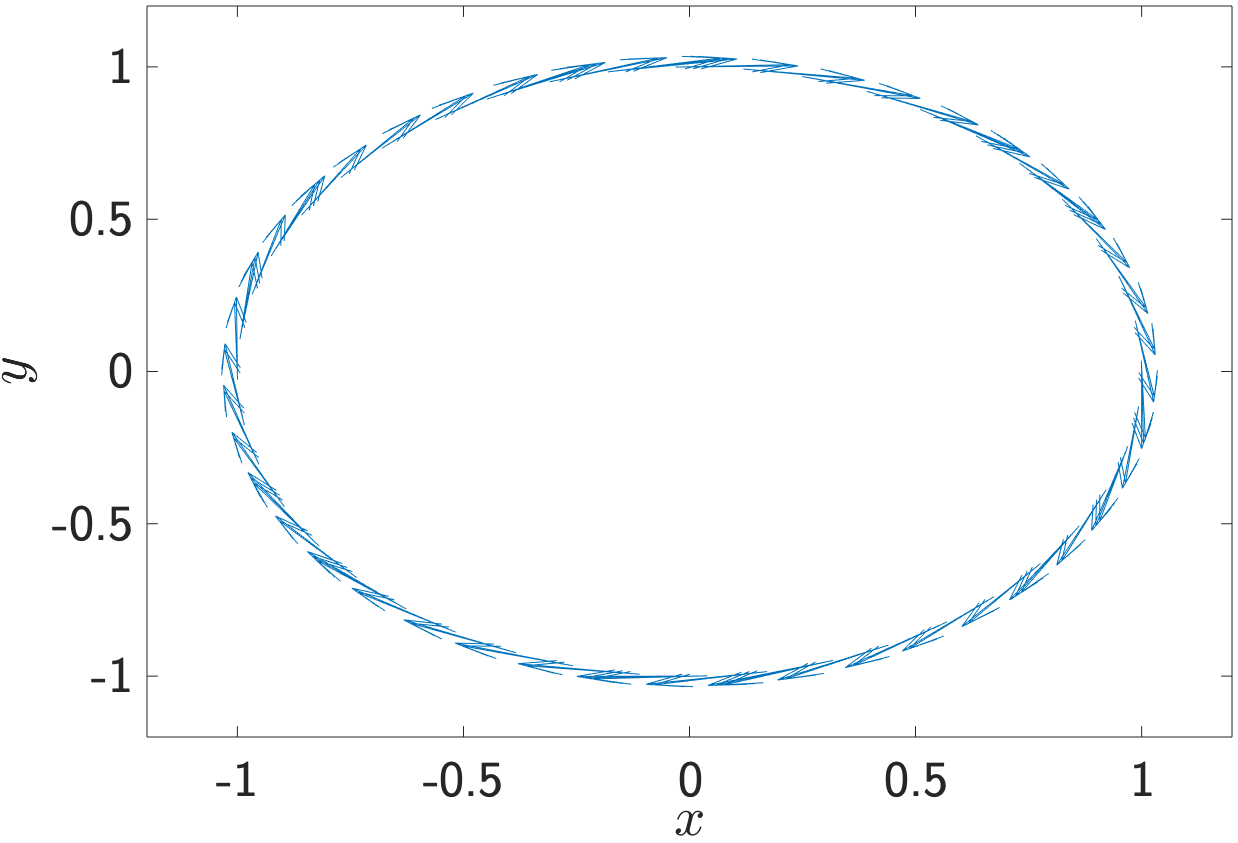}
\caption{\it An example of the limiting circle: Arrows denote the update directions.}\label{fig:limit}
\end{figure}

Limiting cycle is a well-known issue for bilevel machine learning problems [4,5]. The reason behind limiting cycle is that different from minimization problems, a bilevel optimization problem is more complicated and could be highly nonconvex-nonconcave, where the inner problem can not be solved exactly. Here we provide a simple bilevel problem example, which is convex-concave, but the iterations still cannot converge due to the inexact solutions. Specifically, we consider the following optimization problem: $$\min_x \max_y f(x,y)=xy.$$ Then at the $t$-th iteration, the update direction will be $(-y_t,x_t)$. If we start from $(1,0)$ with a stepsize of $0.0001$, this update will result in a limiting circle: $x^2+y^2=1$ and never reach the stable equilibrium $(0,0)$ as shown in Figure~\ref{fig:limit}. 

%% file: appendix.tex

\section{Attacker Architecture}\label{sec:slim}
\begin{figure}[htb!]
	\begin{center}
		\includegraphics[width = 0.75\textwidth]{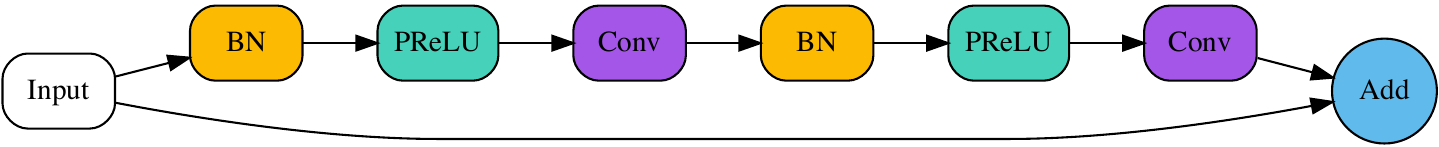}
		\caption{\it An illustration example for the architecture of ResBlocks.}\label{fig:resblocks}
	\end{center}
\end{figure}

In the following, we study how the attacker architecture affects the stability of L2L training.
Table~\ref{tab:optimizer_slim} presents another attacker architecture: \textit{slim attacker}. In this network, the second convolutional layer uses downsampling, while the second last deconvolutional layer uses upsampling. Such a bottleneck design is widely used in deep neural networks due to computational considerations. For example the running time of per epoch for L2L with slim attacker is 480; whereas L2L with the original architecture is 620. However, it loses some information of input and is significant worse than the original architecture (Table~\ref{tab:optimizer}). Inspired by residual learning in~\cite{he2016deep}, we address the stability issue by using a skip layer connection to ease the training of this network. Specifically, the last layer takes the concatenation of $\cA_{f_{\boldsymbol{\theta}}}(\bx,y)$ and the output of the second last layer as input. Figure~\ref{fig:resblocks} presents the architecture of ResBlocks. PReLU is a special type of Leaky ReLU with a learnable slope parameter. 

Table~\ref{tab:result:slim} shows the results of L2L with the slim attacker shown in Table~\ref{tab:optimizer_slim}. 
The performance of GradL2L under PGM attacker on CIFAR10  for slim attacker is comparable to the original attacker. However, under other scenarios, the robust performance is worse than the original attacker. We tried to make the slim attacker deeper or take more L2L steps and observe little improvement. These results suggest a very important design choice of attacker architecture for L2L that the widely used bottleneck design causes the loss of information and can make the training difficult.

\begin{table}[htb!]
	\begin{center}
		\caption{\it Slim Attacker Network Architecture.}\label{tab:optimizer_slim}
		{\normalsize
			\begin{tabular}{ l@{ }l@{ }} 
				\hline
				\hline
				Conv:& [$k = 3 \times 3, c=128, s = 1, p = 1$], BN+ReLU \\
				\hline
				ResBlocks:& [channel = 256]\\
				\hline
				ResBlocks:& [channel = 128],~~BN \\
				\hline
				DeConv:& [$k = 4 \times 4, c=16, s = 2, p = 1$], BN+ReLU \\
				\hline
				Conv:& [$k = 3 \times 3, c=3, s = 1, p = 1$], $\mathrm{tanh}$ \\			
				\hline
				\hline
			\end{tabular}
		}
	\end{center}
\end{table}

\begin{table}[htb!]
	\normalsize
	\begin{center}
		\caption{\it Results of L2L with Slim Attacker under White-box Setting over CIFAR.}\label{tab:result:slim}
		\begin{tabular}{c|c|c|c}
			\hline
			\hline 
			\multirow{2}{*}{Defense Method} &  \multirow{2}{*}{Attack Method}   & \multicolumn{2}{c}{Accuracy}\\
			\cline{3-4}
			& &Clean & Robust
			\\
			\hline\hline \multicolumn{4}{l}{}  \\ [-1.8ex]
			\multicolumn{4}{c}{Dataset: CIFAR10} \\
			\hline
			Grad L2L &    PGM-20       & 85.31\%&53.02\% \\[0.005cm]
			2-Step L2L  &    PGM-20      &  75.36\%&46.12\%\\[0.005cm]
			Grad L2L 	  &    CW      &85.31\% &42.72\% \\[0.005cm]
			2-Step L2L&    CW      &75.36\% & 40.82\%\\[0.005cm]
			\hline\hline \multicolumn{4}{l}{}  \\ [-1.8ex]
			\multicolumn{4}{c}{Dataset: CIFAR100} \\
			\hline
			Grad L2L &    PGM-20       & 60.60\%&27.37\% \\[0.005cm]
			2-Step L2L  &    PGM-20      &  60.23\%&20.23\%\\[0.005cm]
			Grad L2L  &    CW      & 60.60\%&22.14\% \\[0.005cm]
			2-Step L2L&    CW      & 60.23\%& 22.70\%\\[0.005cm]
			\hline
			\hline
		\end{tabular}
	\end{center}
\end{table}

\section{Black-box Attack}\label{app:black}

Under the black-box setting, we first train a surrogate model with the same architecture of the target model but a different random seed, and then attackers generate adversarial examples to attack the target model by querying gradients from the surrogate model.

The black-box attack highly relies on the transferability, which is the property that the adversarial examples of one model are likely to fool others. However, the transferred attack is very unstable, and often has a large variation in its effectiveness. Therefore, results of the black-box setting might not be reliable and effective. Thus we only present one result here to demonstrate the robustness of different models. 

\begin{table}[htb!]
{
	\begin{center}
		\caption{\it Results of the Black-box Setting over CIFAR-10. We Evaluate L2L Methods with Slim Attacker Networks.}\label{tab:cifar_black_small}
		\begin{tabular}{ l@{ }|P{.45in}P{.45in}P{.45in}P{.45in}P{.45in}c} 
			\hline
			\hline
			Surrogate& \multicolumn{2}{c}{Plain Net}&  \multicolumn{2}{c}{FGSM Net}&  \multicolumn{2}{c}{PGM Net}\\
			& FGSM&  PGM10&  FGSM& PGM10&  FGSM& PGM10\\
			\hline	
Plain Net&	40.03&	 5.60&	74.42  &75.25&	67.37&	65.92\\
	FGSM Net&	79.20&	85.02&	\textbf{89.90}&	80.40&	64.28&	63.89\\
	PGM Net&	83.80&	84.73&	84.33&	85.29&	67.05&	65.54\\
	Naive L2L& 45.52&	25.95&	83.99&	77.94&	68.14&	67.13\\
		Grad L2L &	 \textbf{86.10}&	86.87&	87.93&	\textbf{88.01}&	\textbf{71.15}&	\textbf{69.95}\\
	2-Step L2L & 85.83 & \textbf{87.10} & 86.51 & 87.60 & 70.58 & 69.38 \\			\hline
			\hline
		\end{tabular}
	\end{center}
	}

\end{table}

\begin{table}[htb!]
	\begin{center}
		\caption{\it Experiments under the Black-box Setting over CIFAR-100. Note that here We only Evaluate L2L Methods Using the Slim Attacker Network.}\label{tab:cifar_black}
		\begin{tabular}{ l@{ }|P{.45in}P{.45in}P{.45in}P{.45in}P{.45in}c} 
			\hline
			\hline
			Surrogate& \multicolumn{2}{c}{Plain Net}&  \multicolumn{2}{c}{FGSM Net}&  \multicolumn{2}{c}{PGM Net}\\
			& FGSM&  PGM10&  FGSM& PGM10&  FGSM& PGM10\\
			\hline	
			Plain Net		& 21.04&  9.04& 50.57& 54.06& 40.06& 41.30\\
			FGSM Net		& 42.87& 50.73& \textbf{61.68}& 44.70& 39.34& 40.08\\
			PGM Net			& 56.63& 58.34& 56.99& 57.97& 40.19& 39.87\\
			Naive L2L    	& 20.97& 10.47& 50.36& 54.07& 38.63& 39.91\\
			Grad L2L    	& 57.63& 59.62& 59.18& \textbf{61.26}& 41.71& 41.15\\
			2-Step L2L 		&\textbf{58.66} &\textbf{59.31}&58.92& 59.46&\textbf{45.80}& \textbf{45.31}\\
			\hline
			\hline
		\end{tabular}
	\end{center}
\end{table}

%% file: appendix_advcheck.tex
\section{Robustness Evaluation Checklist}\label{app:advcheck}

Recently, there are many works on robustness defense that have been proven ineffective \cite{athalye2018obfuscated,carlini2019evaluating}. Our work follows the most reliable and widely used robust model approach — adversarial training, which finds a set parameters to make the model robust. We do not make any modification to final classifier model. Unlike previous works (e.g., Defense-GAN, \cite{samangouei2018defense}), our model does not take the attacker as a part of the final model and does not use shattered/obfuscated/masked gradient as a defense mechanism. We also demonstrate that the evaluation of the robustness of our proposed L2L method is trustworthy by verifying all items listed in \cite{carlini2019evaluating}.

\subsection{Shattered/Obfuscated/Masked Gradient}
 In this section we verify that our proposed L2L method does not fall into the pitfall of shattered/obfuscated/masked gradient, which have proven ineffective. To see this, we checked every item recommended in Section 3.1 of \cite{athalye2018obfuscated}:
\begin{itemize}
	\item  One-step attacks perform better than iterative attacks:  Figure~\ref{fig:acc_vs_iter_eps} shows that the PGM attack is stronger with larger number of iterations.
	\item Black-box attacks are better than white-box attacks: Appendix~\ref{app:black} shows that the black-box transfer attack is much weaker than white white-box attacks.
	\item Unbounded attacks do not reach $100\%$ success: We evaluate the model robustness against attack with extremely large perturbation to show that unbounded attacks do reach $100\%$ success. Specifically, we use the PGM-10 attack with various perturbation magnitudes $\epsilon \in [0,1]$ and stepsize $\frac{\epsilon}{10}$. Figure~\ref{fig:large_eps} shows that the PGM attack eventually reach $100\%$ success as the perturbation magnitude increases.
\begin{figure}[!htb]
	\centering
	\includegraphics[width = 0.54\textwidth]{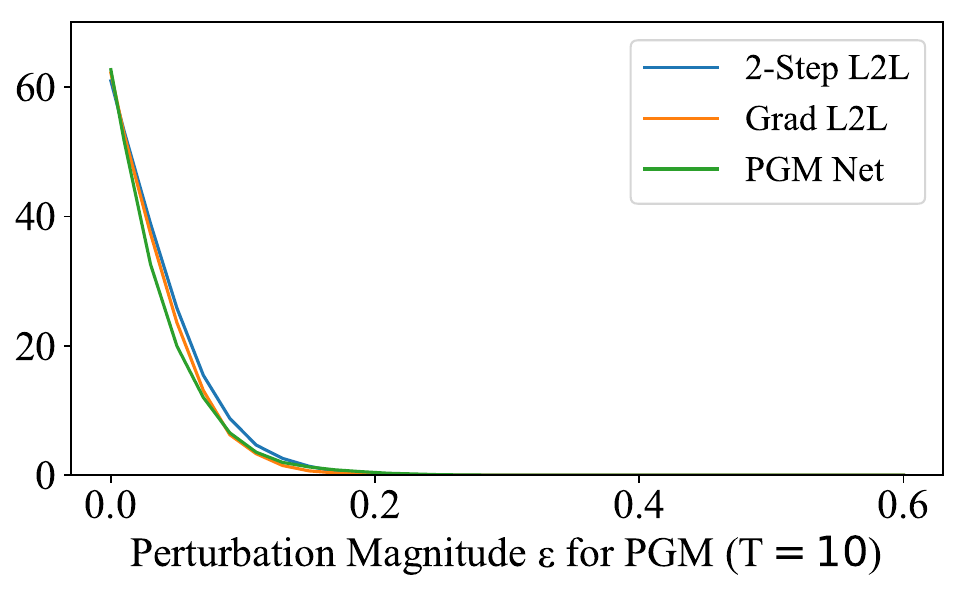}
	\caption{\it Robust accuracy against perturbation magnitudes of PGM over CIFAR-100.}\label{fig:large_eps}
\end{figure}

	\item Random sampling finds adversarial examples: In Table~\ref{tab:cifar_white}, we show that random search is not better than gradient-based method and is rather weak against our model. 
	\item Increasing distortion bound does not increase success: Figure~\ref{fig:acc_vs_iter_eps} shows that the PGM attack becomes stronger as the perturbation magnitude increases.
\end{itemize}

\subsection{Robustness Evaluation Checklist}

\cite{carlini2019evaluating} also provide an evaluation checklist, and we now check each of common severe flaws and common pitfalls as follows:

\begin{itemize}
\item State a precise threat model: We do not have any adversary detector; We do not use shattered/Obfuscated/Masked gradient. We do not have a denoiser. Our model has no aware of the attack mechanism, including PGM and CW attacks. 
\item Adaptive attacks: We used CW, PGM, and L2L attacker attack.
\item Report clean model accuracy: We reported.
\item Do not use Fast Gradient Sign Method. We use PGM-20 and PGM-100 and CW.
\item Do not only use attacks during testing that were used during training. We use different evaluation criteria to evaluate all models.
\item Perform basic sanity tests: It is provided in Figure~\ref{fig:acc_vs_iter_eps}.
\item Generate an attack success rate vs. perturbation budget: Figure~\ref{fig:acc_vs_iter_eps}.
 \item Verify adaptive attacks perform better than any other (e.g., blackbox, and brute-force search):  The above table and Appendix~\ref{app:black} in the paper.
\item Describe the attacks applied: In Section~\ref{sec:exp}.
\item Apply a diverse set of attacks: We tried PGM attack (with different perturbation magnitude and iterations), blackbox attack (transfer attack), CW attack (adaptive attack), L2L attack (adaptive and designed for this particular model), Bruteforce random search (gradient-free attack)
\item  Suggestions for randomized defenses: We are not.
\item  Suggestions for non-differentiable components (e.g., by performing quantization or adding extra randomness): We have no additional non-differentiable component.

\item  Verify that the attacks have converged: Figure~\ref{fig:acc_vs_iter_eps} shows that the PGM attack eventually converges.
\item Carefully investigate attack hyperparameters: Figure~\ref{fig:acc_vs_iter_eps}.
\item Compare against prior work: We compared our algorithm to PGM net. L2L is more computationally efficient and the L2L model is more robust due to the fact that L2L attack is strong enough. Unlike Defense-GAN, we do not use the generator (attacker in L2L) as the denoising module and do not change the final prediction model.
\end{itemize}
